\documentclass{article}

\usepackage[final,nonatbib]{neurips_2025}
\usepackage[numbers,sort&compress]{natbib}

\usepackage[utf8]{inputenc}
\usepackage[T1]{fontenc}
\usepackage{hyperref}
\usepackage{url}
\usepackage{booktabs}
\usepackage{amsfonts}
\usepackage{nicefrac}
\usepackage{microtype}
\usepackage{xcolor}
\usepackage{graphicx}
\usepackage{amsmath,amssymb}
\usepackage{subcaption}
\usepackage{nicematrix}
\usepackage{colortbl}
\usepackage{multirow}
\usepackage{tikz}
\usetikzlibrary{trees}
\usepackage{forest}

\definecolor{maplecolor}{RGB}{255, 0, 0}

\definecolor{darkgreen}{RGB}{24,140,104}
\definecolor{darkred}{RGB}{218,77,56}

\workshoptitle{REO: Advances in Representation Learning for Earth Observation}

\title{MAPLE: Multi-Path Adaptive Propagation with Level-Aware Embeddings for Hierarchical Multi-Label Image Classification}

\author{%
  Boshko Koloski$^{\dagger,1,2}$ \quad
  Marjan Stoimchev$^{\dagger,1,2}$ \\[0.3em]
  Jurica Levatić$^{1}$ \quad
  Dragi Kocev$^{1}$ \quad
  Sašo Džeroski$^{1}$ \\
  $^{1}$ Jo\v{z}ef Stefan Institute, Ljubljana, Slovenia \\
  $^{2}$ Jo\v{z}ef Stefan International Postgraduate School, Ljubljana, Slovenia \\[0.3em]
  $^{\dagger}$These authors contributed equally to this work. \\[0.3em]
  \texttt{\{boshko.koloski, marjan.stoimchev, jurica.levatic\}@ijs.si} \\
  \texttt{\{dragi.kocev, saso.dzeroski\}@ijs.si}
}

\begin{document}

\maketitle

\begin{abstract}
Hierarchical multi-label classification (HMLC) is essential for modeling structured label dependencies in remote sensing. Yet existing approaches struggle in \emph{multi-path} settings, where images may activate multiple taxonomic branches, leading to underuse of hierarchical information. We propose MAPLE (\textit{Multi-Path Adaptive Propagation with Level-Aware Embeddings}), a framework that integrates (i) \emph{hierarchical semantic initialization} from graph-aware textual descriptions, (ii) \emph{graph-based structure encoding} via graph convolutional networks (GCNs), and (iii) \emph{adaptive multi-modal fusion} that dynamically balances semantic priors and visual evidence. An \emph{adaptive level-aware objective} automatically selects appropriate losses per hierarchy level. Evaluations on CORINE-aligned remote sensing datasets (AID, DFC-15, and MLRSNet) show consistent improvements of up to +42\% in few-shot regimes while adding only 2.6\% parameter overhead, demonstrating that MAPLE effectively and efficiently models hierarchical semantics for Earth observation (EO).
\end{abstract}

\let\thefootnote\relax\footnotetext{
\hspace{-1.5em}*\,Correspondence to \texttt{boshko.koloski@ijs.si} and \texttt{marjan.stoimchev@ijs.si}
}

\section{Introduction}
Remote sensing image (RSI) classification demands methods that respect the hierarchical organization of land cover types~\cite{BEN}. While modern deep learning excels at flat multi-label classification (MLC)~\cite{mlc_gcn,resisc45}, it overlooks taxonomic structure in standards like CORINE \cite{corine_landcover}, limiting utility in environmental monitoring, urban planning, and climate assessment.

Hierarchical multi-label classification (HMLC) models label dependencies across levels~\cite{jurica_hmlc,hmlc_networks,use_all_labels}, but faces key limitations: (i) many assume \emph{single-path} hierarchies and fail when images belong to multiple branches; (ii) \emph{network-based} designs \cite{b-cnn,condition-cnn} are computationally heavy while \emph{loss-based} formulations \cite{HRN,c-hmcnn} miss long-range dependencies; and (iii) most remain purely supervised despite prevalent low-label regimes in satellite imagery~\cite{self_supervised_rs_survey,ssl_survey,ssl_survey_v2,safonova2023ten}.

We present \textit{Multi-Path Adaptive Propagation with Level-Aware Embeddings} (MAPLE), a hierarchical multi-label image classification framework for remote sensing that explicitly encodes \emph{multi-path} structure. MAPLE initializes label nodes with \emph{hierarchical semantic embeddings} from contextual templates; fuses these with Vision Transformer (ViT) \cite{vit_original} features via \emph{adaptive multimodal gating}; and refines representations through \emph{graph-based propagation} on the taxonomy. A unified prediction head with an \emph{adaptive level-aware objective} yields consistent predictions across hierarchy levels without manual loss tuning.

\textbf{Contributions.} (i) We introduce a multi-token transformer with graph-based hierarchical reasoning for multi-path HMLC in RSI classification; (ii) we construct CORINE-aligned \emph{multi-path} hierarchies for AID\cite{aid_dataset_new}, DFC-15\cite{dfc_15}, and MLRSNet\cite{mlrsnet}, and evaluate MAPLE across nine datasets spanning remote sensing, medical imaging, and fine-grained visual categorization to assess broader generalizability beyond EO; and (iii) we demonstrate strong performance under limited annotation budgets, achieving up to 42\% gains over flat baselines with only 2.6\% parameter overhead, highlighting MAPLE's accuracy-efficiency balance for EO applications~\cite{safonova2023ten}.

\section{Methodology}

\subsection{Problem Formulation}
Given an input image $\mathbf{x} \in \mathbb{R}^{C \times H \times W}$ and a hierarchical label graph $\mathcal{G} = (\mathcal{V}, \mathcal{E})$, where nodes $\mathcal{V}$ represent labels across $L$ levels and edges $\mathcal{E}$ encode parent-child relationships, our goal is to predict a multi-label vector $\hat{\mathbf{y}} \in \{0,1\}^{|\mathcal{V}|}$ that respects the hierarchical constraints.

\begin{figure}[t]
    \centering
    \includegraphics[width=0.7\textwidth]{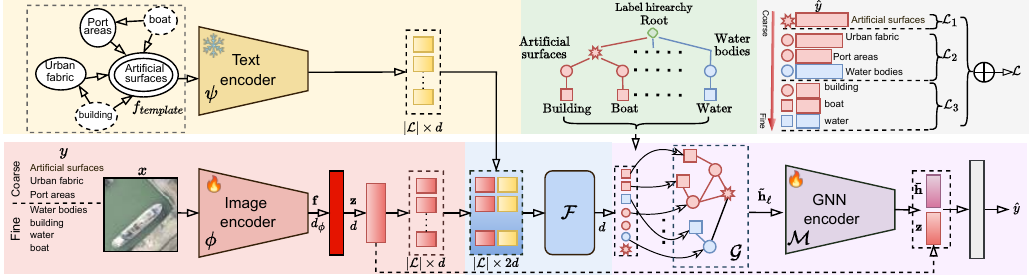}
    \caption{MAPLE architecture overview. The framework processes an input image with a ViT encoder that uses hierarchy-specific class tokens. A GCN refines these token embeddings by propagating information along the label taxonomy. Finally, visual features and refined semantic embeddings are fused via adaptive gating to produce level-aware classifications.}
    \label{fig:method}
\end{figure}

\subsection{Hierarchical Semantic Initialization}

Each node $\ell \in \mathcal{L}$ is initialized with a semantic embedding $\mathbf{e}_\ell^{(0)} \in \mathbb{R}^d$ derived from both its label and its position in the hierarchy. Instead of random initialization, we generate a contextual description $\tau(\ell)$ that combines the node name with its parent and child relations, encoded through a pre-trained sentence transformer $\psi$ and linearly projected to the model dimension:
\[
\mathbf{e}_\ell^{(0)} = \mathrm{norm}\!\left(\mathbf{W}_\psi\, \psi(\tau(\ell))\right)
\]
where $\mathbf{W}_\psi$ is a learnable projection and $\mathrm{norm}(\cdot)$ denotes L2 normalization. This initialization provides semantically meaningful embeddings that reflect the hierarchical organization of the taxonomy. A detailed example is shown in Appendix~\ref{app:implementation}.

\subsection{Visual Representation and Multi-Token Transformer}
We extend the standard ViT by introducing $M=|\mathcal{V}|$ learnable class tokens, one for each node in the hierarchy. These tokens $T_{\text{CLS}} \in \mathbb{R}^{M \times d}$ are prepended to image patch tokens $T_p$, forming the input sequence $T = [T_{\text{CLS}} \| T_p]$. This multi-token design allows each class token to learn specialized, label-specific patterns by attending to relevant image regions. The ViT processes the input through self-attention mechanisms, extracting the CLS token representation as the global image descriptor. Visual features are projected into a common $d$-dimensional latent space through a learnable linear transformation to enable effective multimodal fusion with semantic embeddings.

\subsection{Graph-Based Hierarchical Refinement}
To explicitly leverage the label hierarchy, we refine the class token embeddings using a graph neural network (GNN). The output tokens from the ViT encoder serve as initial node features for a GraphSAGE-style \cite{hamilton2017inductive} message passing process on the graph $\mathcal{G}$. This refinement is performed iteratively for $L_g$ layers:
\begin{equation}
\mathbf{H}^{(k+1)} = \text{GELU}\left(\text{LayerNorm}\left(\mathcal{M}^{(k)}(\mathbf{H}^{(k)}, \mathcal{E}) + \mathbf{H}^{(k)}\right)\right),
\end{equation}
where $\mathcal{M}^{(k)}$ is the message passing function that aggregates information from neighboring nodes. The residual connection preserves original node information while learning relational context. This process allows parent nodes to aggregate features from their children and vice versa, creating robust, hierarchy-aware embeddings.

\subsection{Adaptive Multimodal Fusion}
To produce the final representations, we combine visual information with semantic, hierarchy-aware information through an adaptive gating mechanism. The visual representation is expanded to match all hierarchy nodes by replication. For each node $v$, we compute fusion weights $\boldsymbol{\gamma}_v \in [0,1]^d$ that dynamically balance the contribution of the visual representation $\mathbf{z} \in \mathbb{R}^d$ and the refined token embedding $\mathbf{e}_v \in \mathbb{R}^d$:
\begin{equation}
    \tilde{\mathbf{h}}_v = \boldsymbol{\gamma}_v \odot \mathbf{e}_v + (1 - \boldsymbol{\gamma}_v) \odot \mathbf{z},
\end{equation}
where $\odot$ denotes element-wise multiplication. The fused representation $\tilde{\mathbf{h}}_v \in \mathbb{R}^d$ allows the model to decide, on a per-label basis, whether to rely more on visual cues or learned taxonomic context. The gating weights are computed through a learned network with LayerNorm and sigmoid activation, enabling node-specific fusion strategies across hierarchical levels.

\subsection{Unified Prediction Head and Training Objective}
After graph refinement, node embeddings are mean-pooled and concatenated with the original visual features. A single linear transformation maps this combined representation to classification logits for all hierarchy nodes simultaneously. The output logits are partitioned by level to enable level-specific training.

To accommodate varying label structures at different hierarchy depths, we use an adaptive loss function. At each level $t$, we apply Cross-Entropy (CE) loss if the ground-truth vector $\mathbf{y}_t$ is single-label ($\|\mathbf{y}_t\|_1 = 1$) and Binary Cross-Entropy (BCE) loss otherwise. The total loss averages across all $L$ levels: $\mathcal{L} = \frac{1}{L} \sum_{t=1}^{L} \mathcal{L}_{\text{adaptive}}$. This formulation enables simultaneous supervision across all semantic resolutions while automatically adapting to label distribution characteristics at each hierarchical level.

\section{Experiments}

We evaluate MAPLE on three remote sensing datasets (AID, DFC-15, and MLRSNet) with CORINE-aligned hierarchies~\cite{corine_landcover,aid_dataset_new,dfc_15,mlrsnet}. The model employs a ViT-B/16 backbone~\cite{vit_original} pre-trained on ImageNet~\cite{imagenet} and a 2-layer GraphSAGE network~\cite{hamilton2017inductive}. Performance is reported as mean Area Under the Precision-Recall Curve (AU$\overline{\text{PRC}}$) over three runs (computed with \texttt{scikit-learn}~\cite{scikit}). Dataset statistics, hierarchy construction, and implementation details are provided in Appendices~\ref{app:datasets}–\ref{app:implementation}.

\textbf{Hierarchical vs. Flat Classification.} Table~\ref{tab:main_results} shows that MAPLE consistently outperforms the flat MLC baseline (leaf labels only) across all datasets, with gains from 0.56\% (MLRSNet) to 3.61\% (AID). Improvements are strongest on AID, where the moderate dataset size and rich hierarchical structure allow effective exploitation of CORINE relations~\cite{corine_landcover}. On the larger MLRSNet, MAPLE still yields consistent improvements despite the strong baseline.

\begin{table}[h]
\centering
\tiny
\caption{Hierarchical vs. flat classification (AU$\overline{\text{PRC}}$ \%). MAPLE outperforms the flat baseline at all hierarchy levels ($l_1$–$l_3$) and leaf nodes.}
\label{tab:main_results}
\begin{NiceTabular}[color-inside]{lcccccc}
\toprule
 & \multicolumn{4}{c}{\textbf{MAPLE}} & & \\
\cmidrule(lr){2-5}
Dataset & $l_1$ & $l_2$ & $l_3$ & Leaf & \textbf{MLC} & $\Delta$ (\%) \\
\midrule
AID & 95.31 & 94.32 & 84.83 & \textbf{87.25} & 84.21 & +3.6 \\
DFC-15 & 99.62 & 98.33 & 98.37 & \textbf{98.71} & 98.65 & +0.1 \\
MLRSNet & 98.23 & 97.88 & 96.77 & \textbf{96.71} & 96.17 & +0.6 \\
\bottomrule
\end{NiceTabular}
\end{table}

\textbf{Few-Shot Learning.} Table~\ref{tab:fewshot} highlights MAPLE’s robustness under limited supervision. The largest relative gains occur at 4-shot (AID +25.0\%, DFC-15 +6.6\%, MLRSNet +18.5\%), confirming that hierarchical propagation acts as a strong inductive bias and improves label efficiency across datasets.

\begin{table}[h]
\centering
\tiny
\caption{Few-shot performance (AU$\overline{\text{PRC}}$) with K-shot training per class. $\mu \pm \sigma$ over 3 runs. $\Delta$ shows MAPLE’s improvement over the MLC baseline.}
\label{tab:fewshot}
\begin{NiceTabular}[color-inside]{lcccc}
\toprule
Method & 4-shot & 8-shot & 12-shot & 16-shot \\
\midrule
\multicolumn{5}{c}{\textbf{AID}} \\
\cmidrule(lr){2-5}
\textbf{MLC} & 0.286±0.018 & 0.334±0.012 & 0.341±0.025 & 0.310±0.009 \\
\rowcolor{maplecolor!15} \textbf{MAPLE} & \textbf{0.357±0.021} & \textbf{0.371±0.017} & \textbf{0.396±0.039} & \textbf{0.440±0.046} \\
$\Delta$ (\%) & +25.0 & +11.1 & +16.1 & +41.9 \\
\midrule
\multicolumn{5}{c}{\textbf{DFC-15}} \\
\cmidrule(lr){2-5}
\textbf{MLC} & 0.541±0.044 & 0.541±0.012 & 0.790±0.008 & 0.807±0.021 \\
\rowcolor{maplecolor!15} \textbf{MAPLE} & \textbf{0.577±0.058} & \textbf{0.747±0.036} & \textbf{0.751±0.021} & \textbf{0.854±0.012} \\
$\Delta$ (\%) & +6.6 & +38.1 & -4.9 & +5.8 \\
\midrule
\multicolumn{5}{c}{\textbf{MLRSNet}} \\
\cmidrule(lr){2-5}
\textbf{MLC} & 0.528±0.096 & 0.545±0.000 & 0.675±0.031 & 0.743±0.028 \\
\rowcolor{maplecolor!15} \textbf{MAPLE} & \textbf{0.626±0.011} & \textbf{0.752±0.012} & \textbf{0.762±0.008} & \textbf{0.801±0.017} \\
$\Delta$ (\%) & +18.5 & +38.0 & +12.9 & +7.8 \\
\bottomrule
\end{NiceTabular}
\end{table}

\begin{table}[ht!]
\centering
\tiny
\caption{Comparison with state-of-the-art HMLC methods on \emph{leaf-level} AU$\overline{\text{PRC}}$ ($\uparrow$). Best in \textbf{bold}, second-best \underline{underlined}.}

\label{tab:sota_remote}
\begin{NiceTabular}[color-inside]{lccc}
\toprule
\textbf{Method} & \textbf{AID} & \textbf{DFC-15} & \textbf{MLRSNet} \\
\midrule
C-HMCNN~\cite{c-hmcnn} & 0.764 & 0.962 & 0.792 \\
HiMulConE~\cite{use_all_labels} & \underline{0.770} & \underline{0.970} & \underline{0.865} \\
HMI~\cite{hmi} & 0.647 & 0.923 & 0.437 \\
\rowcolor{maplecolor!15} \textbf{MAPLE (Ours)} & \textbf{0.872} & \textbf{0.987} & \textbf{0.967} \\
\bottomrule
\end{NiceTabular}
\end{table}

\textbf{Comparison with State of the Art.} 
Table~\ref{tab:sota_remote} compares MAPLE with representative state-of-the-art HMLC methods, including C-HMCNN~\cite{c-hmcnn}, HiMulConE~\cite{use_all_labels}, and HMI~\cite{hmi}. 
Evaluated at the \emph{leaf} level, MAPLE attains the best AU$\overline{\text{PRC}}$ on all three datasets by a clear margin, confirming the benefits of multi-path propagation and level-aware embeddings.

Beyond EO, MAPLE generalizes effectively to medical imaging and fine-grained categorization, achieving strong gains on complex hierarchies (PadChest +21.9\%, ETHEC +10.4\%)~\cite{padchest,ethec}. Full results and visualizations are included in Appendices~\ref{app:generalization}–\ref{app:ablation}.

\section{Discussion}

MAPLE effectively captures hierarchical dependencies in remote sensing imagery, outperforming flat baselines with only 2.6\% additional parameters and negligible latency (Appendix~\ref{app:efficiency}). Key insights include: (i) consistent gains across datasets (0.56–3.61\%), (ii) up to 42\% improvement in few-shot regimes, confirming taxonomy-driven inductive bias, and (iii) strong generalization to non-EO hierarchies such as PadChest (+21.9\%).  

The results show that explicitly encoding class hierarchies not only improves predictive accuracy but also enhances stability and robustness to label noise. By constraining predictions along valid hierarchical paths, MAPLE mitigates inconsistent label assignments and promotes semantically coherent outputs. This structural regularization effect becomes especially beneficial in low-data settings, where flat classifiers often overfit or ignore fine-grained relations between categories.  

Beyond accuracy, MAPLE offers interpretable predictions by revealing how decisions propagate through the hierarchy, a property valuable for trust and transparency in applications such as environmental monitoring and medical analysis.  

Its efficiency and label efficiency make MAPLE practical for large-scale or operational EO scenarios where annotations are limited. While gains narrow on massive datasets, consistent improvements across all regimes highlight the importance of hierarchical modeling for structured output prediction. Future work will extend MAPLE to semi-supervised and contrastive learning and explore automatic hierarchy discovery to reduce reliance on expert-defined taxonomies.

\section*{Acknowledgments}
We acknowledge the financial support of the Slovenian Research and Innovation Agency (ARIS) through the core research programme P2-0103 (Knowledge Technologies), projects J1-3033, J2-2505, J2-4452, J2-4660, J3-3070, J4-3095, J5-4575, J7-4636, J7-4637, and N2-0236. The work of the BK was supported by the Young Researcher grant PR-12394. 


\bibliographystyle{unsrtnat}  
\bibliography{references}     

@misc{koloski2025llmembeddingsdeeplearning,
      title={LLM Embeddings for Deep Learning on Tabular Data}, 
      author={Boshko Koloski and Andrei Margeloiu and Xiangjian Jiang and Blaž Škrlj and Nikola Simidjievski and Mateja Jamnik},
      year={2025},
      eprint={2502.11596},
      archivePrefix={arXiv},
      primaryClass={cs.LG},
      url={https://arxiv.org/abs/2502.11596}, 
}

@article{safonova2023ten,
  title={Ten deep learning techniques to address small data problems with remote sensing},
  author={Safonova, Anastasiia and Ghazaryan, Gohar and Stiller, Stefan and Main-Knorn, Magdalena and Nendel, Claas and Ryo, Masahiro},
  journal={International Journal of Applied Earth Observation and Geoinformation},
  volume={125},
  pages={103569},
  year={2023},
  publisher={Elsevier}
}

@ARTICLE{self_supervised_rs_survey,
  author={Wang, Yi and Albrecht, Conrad M. and Braham, Nassim Ait Ali and Mou, Lichao and Zhu, Xiao Xiang},
  journal={IEEE Geoscience and Remote Sensing Magazine}, 
  title={Self-Supervised Learning in Remote Sensing: A review}, 
  year={2022},
  volume={10},
  number={4},
  pages={213-247}}

@article{ssl_survey,
  title={A survey on semi-supervised learning},
  author={Van Engelen, Jesper E and Hoos, Holger H},
  journal={Machine learning},
  volume={109},
  number={2},
  pages={373--440},
  year={2020},
  publisher={Springer}
}

@article{aid_dataset_new,
  title={{R}elation {N}etwork for {M}ulti-label {A}erial {I}mage {C}lassification},
  author={Y. Hua and L. Mou and X.X. Zhu},
  journal={IEEE Transactions on Geoscience and Remote Sensing},
  year={2019}
}

@article{dfc_15,
  title={{R}ecurrently exploring class-wise attention in a hybrid convolutional and bidirectional {LSTM} network for multi-label aerial image classification},
  author={Y. Hua and L. Mou and X.X. Zhu},
  journal={ISPRS Journal of Photogrammetry and Remote Sensing},
  volume={149},
  pages={188--199},
  year={2019},
}

@article{mlrsnet,
    title = {{MLRSN}et: {A} multi-label high spatial resolution remote sensing dataset for semantic scene understanding},
    journal = {ISPRS Journal of Photogrammetry and Remote Sensing},
    volume = {169},
    pages = {337-350},
    year = {2020},
    author = {Q. Xiaoman Qi and Z. Panpan and W. Yuebin and Z. Liqiang and P. Junhuan and W. Mengfan and C. Jialong and Z. Xudong and Z. Ning and P.M.Takis}
}

@ARTICLE{resisc45,
  author={Cheng, Gong and Han, Junwei and Lu, Xiaoqiang},
  journal={Proceedings of the IEEE}, 
  title={Remote Sensing Image Scene Classification: Benchmark and State of the Art}, 
  year={2017},
  volume={105},
  number={10},
  pages={1865-1883}}

@ARTICLE{BEN,
    author = {G. Sumbul and M. Charfuelan and B. Demir and V. Markl},
    title = {{B}ig{E}arth{N}et: {A} {L}arge-{S}cale {B}enchmark {A}rchive for {R}emote {S}ensing {I}mage {U}nderstanding},
    journal = {IEEE International Geoscience and Remote Sensing Symposium},
    volume = {12},
    number = {2},
    pages = {5901--5904},
    year = 2019,
    address = {Berlin}
}

@inproceedings{
vit_original,
title={An Image is Worth 16x16 Words: Transformers for Image Recognition at Scale},
author={Alexey Dosovitskiy and Lucas Beyer and Alexander Kolesnikov and Dirk Weissenborn and Xiaohua Zhai and Thomas Unterthiner and Mostafa Dehghani and Matthias Minderer and Georg Heigold and Sylvain Gelly and Jakob Uszkoreit and Neil Houlsby},
booktitle={International Conference on Learning Representations},
year={2021},
url={https://openreview.net/forum?id=YicbFdNTTy}
}

@INPROCEEDINGS{ImageNet,
  author={J. Deng and W. Dong and R. Socher and L.J. Li and L. Kai and L. Fei-Fei},
  booktitle={IEEE Conference on Computer Vision and Pattern Recognition (CVPR)}, 
  title={{ImageNet: A Large-Scale Hierarchical Image Database}}, 
  year={2009},
  volume={},
  number={},
  pages={248-255},}

@article{scikit,
 title={Scikit-learn: Machine Learning in {P}ython},
 author={Pedregosa, F. and Varoquaux, G. and Gramfort, A. and Michel, V.
         and Thirion, B. and Grisel, O. and Blondel, M. and Prettenhofer, P.
         and Weiss, R. and Dubourg, V. and Vanderplas, J. and Passos, A. and
         Cournapeau, D. and Brucher, M. and Perrot, M. and Duchesnay, E.},
 journal={Journal of Machine Learning Research},
 volume={12},
 pages={2825--2830},
 year={2011}
}

@article{padchest,
title = {PadChest: A large chest x-ray image dataset with multi-label annotated reports},
journal = {Medical Image Analysis},
volume = {66},
pages = {101797},
year = {2020},
issn = {1361-8415},
author = {Aurelia Bustos and Antonio Pertusa and Jose-Maria Salinas and Maria {de la Iglesia-Vayá}},
}

@article{hpa,
  title={Analysis of the human protein atlas image classification competition},
  author={Ouyang, Wei and Winsnes, Casper F and Hjelmare, Martin and Cesnik, Anthony J and {\AA}kesson, Lovisa and Xu, Hao and Sullivan, Devin P and Dai, Shubin and Lan, Jun and Jinmo, Park and others},
  journal={Nature methods},
  volume={16},
  number={12},
  pages={1254--1261},
  year={2019},
  publisher={Nature Publishing Group US New York}
}

@misc{hpa_kaggle,
  author = {Winsnes, Casper and Sullivan, Devin and Park, Elizabeth and Lundberg, Emma and Hjelmare, Martin and Culliton, Phil},
  title = {Human Protein Atlas Image Classification},
  year = {2018},
  note = {Kaggle Competition},
  url = {https://kaggle.com/competitions/human-protein-atlas-image-classification}
}

@article{mured,
  title={Multi-label retinal disease classification using transformers},
  author={Rodr{\'\i}guez, Manuel Alejandro and AlMarzouqi, Hasan and Liatsis, Panos},
  journal={IEEE Journal of Biomedical and Health Informatics},
  volume={27},
  number={6},
  pages={2739--2750},
  year={2022},
  publisher={IEEE}
}

@article{corine_landcover,
  title={Corine land cover},
  author={Copernicus, Referans},
  journal={Copernicus Land Monitoring Service. L. Monit. Serv},
  year={2018}
}

@article{icd10,
  title={The ICD-10 classification of mental and behavioural disorders: clinical descriptions and diagnostic guidelines},
  author={World Health Organization and others},
  journal={World Health Organization},
  volume={362},
  year={1992}
}

@article{jurica_hmlc,
  author    = {Jurica Levati\'{c} and Dragi Kocev and Sa\v{s}o D\v{z}eroski},
  title     = {The importance of the label hierarchy in hierarchical multi-label classification},
  journal   = {Journal of Intelligent Information Systems},
  year      = {2015},
  volume    = {45},
  number    = {2},
  pages     = {247--271},
  month     = {October},
  issn      = {1573-7675}
}

@article{condition-cnn,
  title={Condition-{CNN}: A hierarchical multi-label fashion image classification model},
  author={Kolisnik, Brendan and Hogan, Isaac and Zulkernine, Farhana},
  journal={Expert Systems with Applications},
  volume={182},
  pages={115195},
  year={2021},
  publisher={Elsevier}
}

@InProceedings{use_all_labels,
    author    = {Zhang, Shu and Xu, Ran and Xiong, Caiming and Ramaiah, Chetan},
    title     = {Use All the Labels: A Hierarchical Multi-Label Contrastive Learning Framework},
    booktitle = {Proceedings of the IEEE/CVF Conference on Computer Vision and Pattern Recognition (CVPR)},
    month     = {June},
    year      = {2022},
    pages     = {16660-16669}
}

@article{b-cnn,
  title={B-CNN: branch convolutional neural network for hierarchical classification},
  author={Zhu, Xinqi and Bain, Michael},
  journal={arXiv preprint arXiv:1709.09890},
  year={2017}
}

@inproceedings{HRN,
  title={Label relation graphs enhanced hierarchical residual network for hierarchical multi-granularity classification},
  author={Chen, Jingzhou and Wang, Peng and Liu, Jian and Qian, Yuntao},
  booktitle={Proceedings of the IEEE/CVF Conference on Computer Vision and Pattern Recognition},
  pages={4858--4867},
  year={2022}
}

@article{c-hmcnn,
  title={Coherent hierarchical multi-label classification networks},
  author={Giunchiglia, Eleonora and Lukasiewicz, Thomas},
  journal={Advances in neural information processing systems},
  volume={33},
  pages={9662--9673},
  year={2020}
}

@inproceedings{mlc_gcn,
  title={Multi-label image recognition with graph convolutional networks},
  author={Chen, Zhao-Min and Wei, Xiu-Shen and Wang, Peng and Guo, Yanwen},
  booktitle={Proceedings of the IEEE/CVF conference on computer vision and pattern recognition},
  pages={5177--5186},
  year={2019}
}

@inproceedings{hmlc_networks,
  title={Hierarchical multi-label classification networks},
  author={Wehrmann, Jonatas and Cerri, Ricardo and Barros, Rodrigo},
  booktitle={International conference on machine learning},
  pages={5075--5084},
  year={2018},
  organization={PMLR}
}

@inproceedings{hmi,
 author = {Xiong, Bo and Cochez, Michael and Nayyeri, Mojtaba and Staab, Steffen},
 booktitle = {Advances in Neural Information Processing Systems},
 editor = {S. Koyejo and S. Mohamed and A. Agarwal and D. Belgrave and K. Cho and A. Oh},
 pages = {33016--33028},
 publisher = {Curran Associates, Inc.},
 title = {Hyperbolic Embedding Inference for Structured Multi-Label Prediction},
 volume = {35},
 year = {2022}
}

@article{ssl_survey_v2,
  title={A survey on deep semi-supervised learning},
  author={Yang, Xiangli and Song, Zixing and King, Irwin and Xu, Zenglin},
  journal={IEEE Transactions on Knowledge and Data Engineering},
  volume={35},
  number={9},
  pages={8934--8954},
  year={2022},
  publisher={IEEE}
}

@inproceedings{hgclip,
  title={HGCLIP: Exploring Vision-Language Models with Graph Representations for Hierarchical Understanding},
  author={Xia, Peng and Yu, Xingtong and Hu, Ming and Ju, Lie and Wang, Zhiyong and Duan, Peibo and Ge, Zongyuan},
  booktitle={Proceedings of the 31st International Conference on Computational Linguistics},
  pages={269--280},
  year={2025}
}

@inproceedings{oxford_pets,
  title={Cats and dogs},
  author={Parkhi, Omkar M and Vedaldi, Andrea and Zisserman, Andrew and Jawahar, CV},
  booktitle={2012 IEEE conference on computer vision and pattern recognition},
  pages={3498--3505},
  year={2012},
  organization={IEEE}
}

@INPROCEEDINGS{stanford_cars,
  author={Krause, Jonathan and Stark, Michael and Deng, Jia and Fei-Fei, Li},
  booktitle={2013 IEEE International Conference on Computer Vision Workshops}, 
  title={3D Object Representations for Fine-Grained Categorization}, 
  year={2013},
  volume={},
  number={},
  pages={554-561}}

@inproceedings{ethec,
  title={Hierarchical image classification using entailment cone embeddings},
  author={Dhall, Ankit and Makarova, Anastasia and Ganea, Octavian and Pavllo, Dario and Greeff, Michael and Krause, Andreas},
  booktitle={Proceedings of the IEEE/CVF conference on computer vision and pattern recognition workshops},
  pages={836--837},
  year={2020}
}

@article{sentence_transformers,
  title={Huggingface's transformers: State-of-the-art natural language processing},
  author={Wolf, Thomas and Debut, Lysandre and Sanh, Victor and Chaumond, Julien and Delangue, Clement and Moi, Anthony and Cistac, Pierric and Rault, Tim and Louf, R{\'e}mi and Funtowicz, Morgan and others},
  journal={arXiv preprint arXiv:1910.03771},
  year={2019}
}

@article{mpnet,
  title={Mpnet: Masked and permuted pre-training for language understanding},
  author={Song, Kaitao and Tan, Xu and Qin, Tao and Lu, Jianfeng and Liu, Tie-Yan},
  journal={Advances in neural information processing systems},
  volume={33},
  pages={16857--16867},
  year={2020}
}

@article{hamilton2017inductive,
  title={Inductive representation learning on large graphs},
  author={Hamilton, Will and Ying, Zhitao and Leskovec, Jure},
  journal={Advances in neural information processing systems},
  volume={30},
  year={2017}
}

@inproceedings{glove,
  title={Glove: Global vectors for word representation},
  author={Pennington, Jeffrey and Socher, Richard and Manning, Christopher D},
  booktitle={Proceedings of the 2014 conference on empirical methods in natural language processing (EMNLP)},
  pages={1532--1543},
  year={2014}
}

@article{ord2vec,
  title={Word2Vec},
  author={Church, Kenneth Ward},
  journal={Natural Language Engineering},
  volume={23},
  number={1},
  pages={155--162},
  year={2017},
  publisher={Cambridge University Press}
}

@article{nvembedv2,
  title={Nv-embed: Improved techniques for training llms as generalist embedding models},
  author={Lee, Chankyu and Roy, Rajarshi and Xu, Mengyao and Raiman, Jonathan and Shoeybi, Mohammad and Catanzaro, Bryan and Ping, Wei},
  journal={arXiv preprint arXiv:2405.17428},
  year={2024}
}

@inproceedings{velivckovic2017graph,
  title={Graph attention networks},
  author={Veli{\v{c}}kovi{\'c}, Petar and Cucurull, Guillem and Casanova, Arantxa and Romero, Adriana and Lio, Pietro and Bengio, Yoshua},
  booktitle={International Conference on Learning Representations (ICLR)},
  year={2018}
}

@inproceedings{kipf2016semi,
author = {Kipf, Thomas and Welling, Max},
year = {2017},
month = {09},
pages = {},
title = {Semi-Supervised Classification with Graph Convolutional Networks},
booktitle={International Conference on Learning Representations (ICLR)},
}

\appendix

\section{Appendix}

\subsection{Datasets and Hierarchy Construction}\label{app:datasets}



To assess the broader applicability of our hierarchical multi-label learning approach beyond EO, we evaluated MAPLE across nine datasets spanning three distinct domains. Our evaluation includes three publicly available RSI datasets: AID \cite{aid_dataset_new}, DFC-15 \cite{dfc_15}, and MLRSNet \cite{mlrsnet}; three medical imaging datasets: MuRed \cite{mured}, HPA \cite{hpa_kaggle, hpa}, and PadChest \cite{padchest}; and three fine-grained visual categorization (FGVC) benchmark datasets: Oxford Pets-37 \cite{oxford_pets}, Stanford Cars \cite{stanford_cars}, and ETHEC \cite{ethec}. These datasets exhibit varying characteristics in terms of their original label structures and hierarchical organization. The remote sensing and medical imaging datasets are inherently multi-label at the leaf level, as images can contain multiple land cover types or medical conditions simultaneously. The FGVC datasets, in contrast, already possess established hierarchical taxonomies that reflect natural categorical relationships within their respective domains.

\begin{figure*}[ht]
\centering

\begin{subfigure}[b]{\textwidth}
\centering
\resizebox{1\textwidth}{!}{%
\begin{forest}
  for tree={
    draw,
    rectangle,
    rounded corners=1pt,
    align=center,
    font=\tiny,
    inner sep=1pt,
    anchor=center,
    l sep=3mm,
    s sep=1mm,
    edge={-},
    parent anchor=south,
    child anchor=north,
    tier/.wrap pgfmath arg={tier #1}{level()},
  }
  [Root
    [Artificial\\surfaces
      [Urban\\fabric
        [buildings]
        [mobile-home]
      ]
      [{Industrial, commercial and transport units}
        [Industrial or\\commercial units
          [tanks]
        ]
        [Road and rail\\networks and\\associated land
          [cars]
          [pavement]
        ]
        [Port areas
          [dock]
          [ship]
        ]
        [Airports
          [airplane]
        ]
      ]
      [{Mine, dump and construction sites}
        [bare-soil]
      ]
      [{Artificial, non-agricultural vegetated areas}
        [Sport and leisure\\facilities
          [court]
        ]
      ]
    ]
    [Agricultural\\areas
      [Arable\\land
        [field]
      ]
    ]
    [Forest and\\semi-natural\\areas
      [Forests
        [trees]
      ]
      [Scrub and/or\\herbaceous\\vegetation\\associations
        [chaparral]
        [grass]
      ]
    ]
    [Water\\bodies
      [Inland\\waters
        [water]
      ]
      [Marine\\waters
        [sand]
        [sea]
      ]
    ]
  ]
\end{forest}%
}
\caption{AID dataset (CORINE Land Cover nomenclature)}
\label{fig:aid_hierarchy}
\end{subfigure}

\vspace{0.5cm}

\begin{subfigure}[b]{\textwidth}
\centering
\resizebox{1\textwidth}{!}{%
\begin{forest}
  for tree={
    draw, rectangle, rounded corners=1pt, align=center, font=\tiny,
    inner sep=1pt, anchor=center, l sep=3mm, s sep=1mm, edge={-},
    parent anchor=south, child anchor=north, tier/.wrap pgfmath arg={tier #1}{level()},
  }
  [Root
    [Retinal\\Disorders
      [Vascular\\Retinopathies
        [DR]
        [DN]
        [BRVO]
        [CRVO]
        [HTR]
      ]
      [Degenerative\\Retinal\\Disorders
        [ARMD]
        [Molecular\\Disorders
          [MH]
          [CSR]
        ]
        [Retinoschisis\\and Related\\Conditions
          [RS]
          [TSLN]
        ]
        [Choroidal\\Disorders
          [CNV]
        ]
      ]
      [Structural\\Retinal\\Changes
        [LS]
        [ASR]
        [CRS]
      ]
    ]
    [Optic Nerve\\and Disc\\Disorders
      [Optic Disc\\Anomalies
        [ODC]
        [ODP]
      ]
      [Optic Disc\\Edema
        [ODE]
      ]
      [Refractive\\Disorders
        [MYA]
      ]
    ]
    [Other Retinal\\and Optic Nerve\\Disorders
      [Rare\\Diseases
        [OTHER]
      ]
    ]
    [Normal\\Retina
      [NORMAL]
    ]
  ]
\end{forest}}
\caption{MuRed dataset (ICD-10 classification)}
\label{fig:mured_hierarchy}
\end{subfigure}

\caption{Examples of constructed label hierarchies for (a) the AID dataset, derived from the CORINE Land Cover (CLC) nomenclature, and (b) the MuRed dataset, with abbreviations in leaf labels corresponding to ICD-10 codes of disease names.}
\label{fig:example_hierarchies}
\end{figure*}
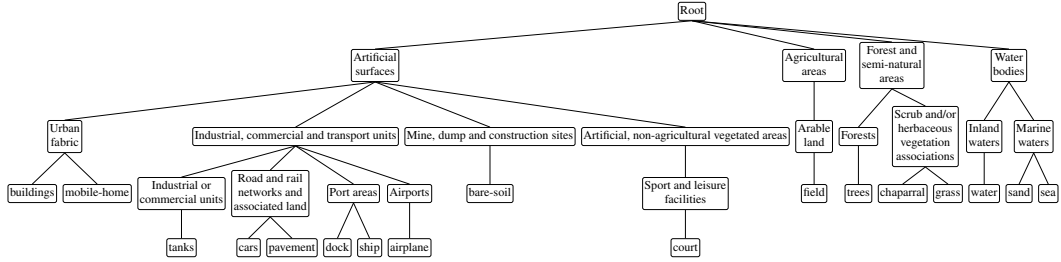
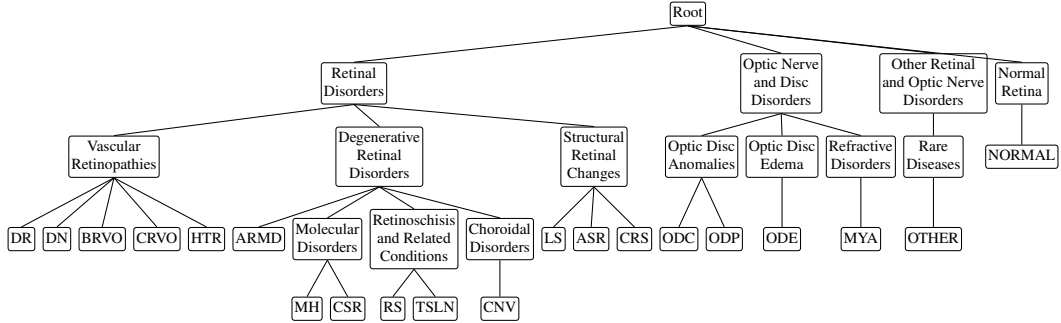

Since comprehensive HMLC image datasets are limited, we adapted these datasets to the hierarchical multi-label setting through domain-appropriate approaches. For the RSI datasets, we constructed hierarchical label structures by systematically mapping the inherently multi-label leaf categories to the CORINE Land Cover (CLC) nomenclature \cite{corine_landcover}. The CLC provides a comprehensive and standardized taxonomy of land cover classes across multiple hierarchical levels, enabling us to create meaningful hierarchical relationships while preserving the multi-label nature at the leaf level. For the medical imaging datasets, we organized the inherently multi-label medical conditions into hierarchical structures based on  the International Classification of Diseases, 10th Revision (ICD-10) codes \cite{icd10}, a clinically validated classification system. This approach ensures that the resulting hierarchies reflect genuine medical taxonomic relationships while maintaining the multi-label characteristics essential for realistic diagnostic scenarios. For the FGVC datasets, we used the existing hierarchical structures inherent to each domain, such as the natural breed taxonomies for Oxford Pets-37 and manufacturer-model relationships for Stanford Cars, which already provide meaningful hierarchical organizations suitable for hierarchical classification tasks.

When direct mapping to established nomenclatures was not feasible due to dataset-specific terminologies or emerging categories, we supplement our approach by querying ChatGPT, followed by a manual inspection, to generate appropriate hierarchical placements \cite{hgclip}. This hybrid strategy ensures comprehensive coverage while maintaining the benefits of standardized taxonomic structures, providing a solid foundation for hierarchical learning across diverse visual recognition tasks. Figure~\ref{fig:example_hierarchies} illustrates representative examples of the constructed hierarchies, demonstrating how the CORINE Land Cover nomenclature and ICD-10 codes translate into structured taxonomic relationships for the remote sensing and medical domain datasets, respectively.

\begin{figure}[t]
   \centering
   \begin{minipage}[b]{0.9\textwidth}
       \centering
       \includegraphics[width=\textwidth]{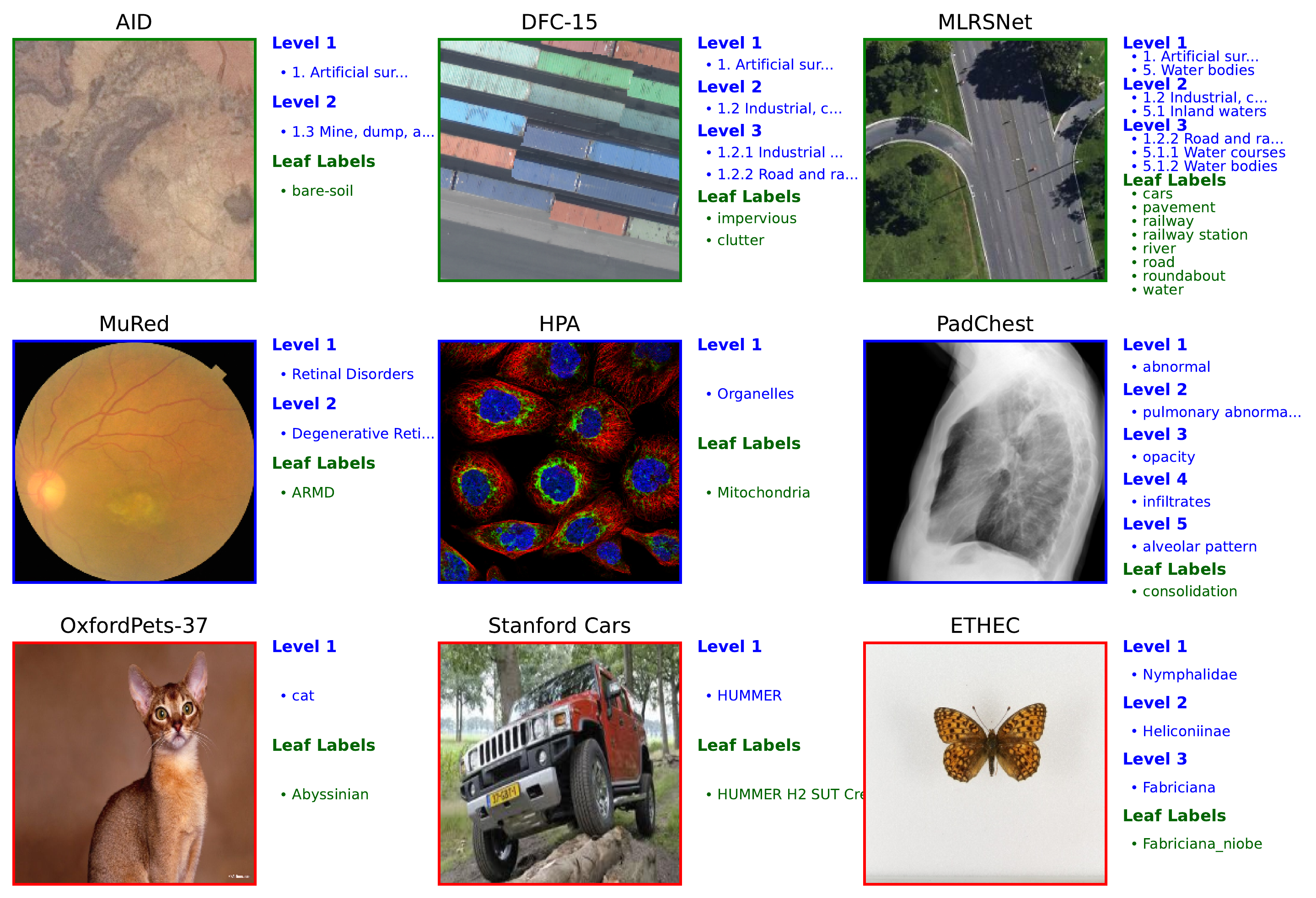}
   \end{minipage}
   \caption{Representative examples from the nine datasets used in our evaluation, showing sample images alongside their corresponding hierarchical label structures. The datasets span three domains: remote sensing (AID, DFC-15, MLRSNet), medical imaging (MuRed, HPA, PadChest), and fine-grained visual categorization (OxfordPets-37, Stanford Cars, ETHEC). Each hierarchy displays the multi-level taxonomic organization from coarse-grained categories at Level 1 to fine-grained leaf labels.}
   \label{fig:dataset_examples}
\end{figure}

Figure~\ref{fig:dataset_examples} provides representative examples from all nine datasets, illustrating the diversity of visual content and the varying complexity of hierarchical structures across domains. The hierarchical organizations range from relatively simple two-level structures (e.g., Oxford Pets-37 and Stanford Cars) to complex multi-level taxonomies with up to six hierarchical levels (e.g., PadChest). Table~\ref{tab:datasets} presents detailed characteristics of all datasets, including the number of labels at each hierarchical level and the dataset splits used for evaluation.

\begin{table}[h]
\centering
\scriptsize
\setlength{\tabcolsep}{2.1pt} 
\caption{Summary of the datasets used in this study. $N$ denotes the total number of images in each dataset, while $N_{train}$ and $N_{test}$ represent the number of images in the training and test sets, respectively. $|\mathcal{L}|$ indicates the number of unique labels at each hierarchical level, where ${\ell}$ corresponds to the deepest (leaf) level.}
\label{tab:datasets}
\begin{NiceTabular}{l l l l|ccccccc}
\toprule
& $N$ & $N_{train}$ & $N_{test}$ & \multicolumn{7}{c}{$|\mathcal{L}|$} \\
Dataset & & & & $1$ & $2$ & $3$ & $4$ & $5$ & $6$ & ${\ell}$ \\
\midrule
AID & 3,000 &  2,400 & 600 & 4 & 9 & 15 & 7 & - & - & 17 \\
DFC-15 & 3,341 &  2,672 &  669& 3 & 7 & 7 & - & - & - & 8 \\
MLRSNet & 109,151 & 87,336 & 21,815 & 7 & 15 & 22 & 60 & - & - & 60 \\
\midrule
MuRed & 2,208 & 1,764 & 444 & 4 & 8 & 17 & 5 & - & - & 20 \\
HPA & 31,072 & 24,837 & 6,235 & 6 & 28 & - & - & - & - & 28 \\
PadChest & 121,230 & 97,203 & 24,027 & 2 & 5 & 9 & 9 & 7 & 2 & 19 \\
\midrule
OxfordPets-37 & 7,349 & 3,680 & 3,669 & 2 & 37 & - & - & - & - & 37 \\
Stanford Cars & 16,185 & 8,144 & 8,041 & 9 & 196 & - & - & - & - & 196 \\
ETHEC & 47,978 & 42,929 & 5,049 & 6 & 21 & 135 & 561 & - & - & 561 \\
\bottomrule
\end{NiceTabular}
\end{table}

\subsubsection{Remote Sensing Image Datasets}

The AID dataset consists of 3,000 aerial images with a resolution of $600 \times 600$ pixels, originally categorized into 30 scene classes for single-label classification \cite{aid_dataset_new}. For our hierarchical multi-label extension, we mapped these scene categories to the CORINE Land Cover nomenclature, resulting in a four-level hierarchy with 35 labels organized across hierarchical levels, providing a structured classification framework for aerial scene understanding. The DFC-15 dataset, originating from the 2015 IEEE GRSS Data Fusion Contest, comprises 3,341 image patches with a resolution of $600 \times 600$ pixels \cite{dfc_15}. Originally designed for semantic segmentation, we adapted it for hierarchical classification by organizing the labels into a three-level hierarchy with 17 distinct labels based on land cover taxonomies. The MLRSNet dataset includes 109,151 images with a resolution of $256 \times 256$ pixels, originally designed for multi-label scene classification with 60 categories \cite{mlrsnet}. The hierarchical version expands this structure to include 104 labels organized across four hierarchical levels using CORINE Land Cover mapping, making it the most complex dataset among the RSI datasets considered in this study.

\subsubsection{Medical Image Datasets}

The Multi-label Retinal Disease (MuRed) dataset \cite{mured} is a publicly available collection of 2,208 retinal fundus images originally designed for MLC of retinal diseases. For our hierarchical extension, we organized the 20 original disease labels using ICD-10 codes\cite{icd10}, resulting in a four-level hierarchy with 34 labels. The Human Protein Atlas (HPA) Image Classification dataset \cite{hpa_kaggle, hpa} contains 28 distinct protein labels and consists of 31,072 samples. The hierarchical extension organizes the 28 protein labels into a two-level hierarchy with 34 total labels based on cellular localization and functional relationships. The PadChest dataset is a large-scale chest X-ray dataset comprising 160,868 images from 67,000 patients, acquired at Hospital San Juan (Spain) between 2009 and 2017 \cite{padchest}. The original dataset contains 193 labels, including 174 radiographic findings, 19 differential diagnoses, and 104 anatomical locations, already organized hierarchically and mapped to the Unified Medical Language System (UMLS). Approximately 27\% of the labels were manually annotated by board-certified radiologists. We utilized a subset of 121,230 samples and organized them into a refined hierarchy of 32 labels across six hierarchical levels based on ICD-10 codes, maintaining the clinical relevance of the original taxonomic organization.

\subsubsection{Fine-Grained Visual Categorization Datasets}

The Oxford Pets-37 dataset comprises 7,349 images of 37 different pet breeds\cite{oxford_pets}. We organized the breeds into a two-level hierarchy based on pet species (cats vs. dogs). The Stanford Cars dataset contains 16,185 images of 196 car models\cite{stanford_cars}. We structured this into a two-level hierarchy based on manufacturer relationships, creating 9 top-level manufacturer categories and 196 specific model categories. The ETHEC dataset, with 47,978 images spanning 561 categories across four hierarchical levels, represents one of the most comprehensive fine-grained datasets with inherent hierarchical structure. Originally designed for hierarchical classification, this dataset provides a natural four-level taxonomy that progresses from broad categorical distinctions to specific sub-categories, offering an extensive evaluation of hierarchical classification capabilities in complex taxonomic structures.

\subsection{Implementation Details}\label{app:implementation}

MAPLE is implemented in \texttt{PyTorch Lightning}\footnote{\url{https://github.com/Lightning-AI/pytorch-lightning}} with a ViT-B/16 backbone (ImageNet initialization). We introduce one learnable class token per label node in the hierarchy, departing from the standard single-token design. This multi-token approach allows each token to specialize in detecting its corresponding semantic category by attending to relevant image regions.

\textbf{Graph-Based Refinement:} A two-layer GraphSAGE network implemented in \texttt{PyTorch Geometric (PyG)}\footnote{\url{https://pytorch-geometric.readthedocs.io/en/latest/}} performs iterative message passing on the label graph. Each layer aggregates information from neighboring nodes (parents and children) in the hierarchy, with residual connections, LayerNorm, and GELU activation ensuring stable training. Dropout (rate=0.1) is applied between layers for regularization.

\textbf{Image Processing:} RGB images resized to $224\times224$ pixels; latent dimension $d{=}768$ matches the ViT-B/16 output dimension.

\textbf{Augmentations:} Random horizontal flips, color jitter (brightness=0.2, contrast=0.2, saturation=0.2), and random resized crops (scale=0.8 to 1.0).

\textbf{Optimization:} AdamW optimizer with base learning rate $1{\times}10^{-4}$, weight decay $1{\times}10^{-2}$, cosine learning rate scheduler with linear warmup (10 epochs), batch size 16, 150 epochs total. Mixed precision training (FP16) is employed to reduce memory consumption and accelerate training.

\paragraph{Hierarchical Semantic Initialization (Sentence Transformer).}
For transformer-based initialization, we employ the \texttt{all-mpnet-base-v2} model from Hugging Face \texttt{Sentence Transformers}~\cite{sentence_transformers, mpnet}. 
For each node in the hierarchy, we construct a short natural-language prompt encoding its taxonomic context:
\begin{quote}\small
\texttt{The category '[label]' which is a subcategory of [parent] and includes subcategories like [child\_1, child\_2, child\_3].}
\end{quote}
If a node has no children, the final clause is omitted. These prompts are encoded into 768-dimensional sentence embeddings, projected to the model dimension $d$ via a learnable linear layer, and L2-normalized to initialize hierarchy-specific class tokens.

Figure~\ref{fig:sentence_transformer_init} illustrates this process for a CORINE-derived branch where only the \emph{ship} class is active. The left panel shows the corresponding path; the right panel presents the instantiated prompts for both the parent and leaf nodes.

\begin{figure}[t]
\centering
\begin{minipage}{0.48\linewidth}
\centering
\scriptsize
\begin{forest}
for tree={
  draw, rectangle, rounded corners=1pt, align=center,
  font=\scriptsize, inner sep=2pt, l sep=5pt, s sep=3pt,
  edge={-}
}
[Root
  [Artificial\\Surfaces
    [Industrial\\Commercial\\and Transport\\Units
      [airplane, text=gray]
      [cars, text=gray]
      [court, text=gray]
      [dock, text=gray]
      [ship, draw, rounded corners=1pt, fill=maplecolor!20]
      [storage\\tanks, text=gray]
    ]
  ]
]
\end{forest}
\end{minipage}\hfill
\begin{minipage}{0.48\linewidth}
\scriptsize
\textbf{Parent node prompt} (\emph{Industrial, Commercial and Transport Units}): \\
\texttt{The category 'Industrial, Commercial and Transport Units' which is a subcategory of Artificial Surfaces and includes subcategories like airplane, cars, court, dock, ship, and storage tanks} \\[6pt]
\textbf{Leaf node prompt} (\emph{ship}): \\
\texttt{The category 'ship' which is a subcategory of Industrial, Commercial and Transport Units.} \\[6pt]
\textbf{Embedding:} $\mathbf{e}_\ell^{(0)} = \mathrm{norm}\!\big(\mathbf{W}_\psi\,\psi(\tau(\ell))\big)$
\end{minipage}
\caption{Hierarchical semantic initialization using the Sentence Transformer on a CORINE-derived path. Left: subgraph with the active node (\emph{ship}) highlighted. Right: instantiated prompts for parent and leaf nodes used for semantic embedding generation.}
\label{fig:sentence_transformer_init}
\end{figure}

\textbf{Adaptive Multimodal Fusion:} Visual features from the ViT encoder are replicated for all nodes and concatenated with semantic embeddings. A learned gating network (linear layer + LayerNorm + Sigmoid) computes per-node, per-dimension fusion weights, allowing dynamic balancing between semantic priors and visual evidence.

\textbf{Unified Prediction Head:} After GNN refinement, node embeddings are mean-pooled and concatenated with the original visual features. A single linear layer maps this combined representation to logits for all hierarchy nodes simultaneously. The output is partitioned by level for loss computation.

\subsection{Evaluation Strategy}\label{app:evaluation}

We train on fixed splits (Table~\ref{tab:datasets}) and report metrics on the test set. For FGVC datasets (Oxford Pets-37, Stanford Cars, ETHEC), we use official train/test splits. For remote sensing and medical datasets, we use 80/20 train/test splits with iterative stratification to preserve label frequency distributions. A validation set (10\% of training data) is used for early stopping and hyperparameter selection.

Unless stated otherwise, performance is reported on leaf labels only to enable fair comparison with flat baselines that predict only at the most specific level. However, MAPLE predicts at all hierarchy levels, and we report per-level performance in detailed analyses.

\textbf{Few-Shot Settings:} To study label scarcity effects, we conduct few-shot experiments with $K\in\{4,8,12,16\}$ labeled examples per leaf category. For each shot configuration, we randomly sample $K$ examples per leaf class from the training set, ensuring balanced representation. We perform three independent runs with different random seeds and report mean and standard deviation of AU$\overline{\textrm{PRC}}$ on the test set.

\subsection{Computational Resources}\label{app:resources}

All experiments were conducted on four NVIDIA A100 GPUs equipped with 40 GB memory each.

\subsubsection{Evaluation Metrics}\label{app:metrics}

We employ the micro-averaged Area Under the Precision-Recall Curve  (AU$\overline{\text{PRC}}$), a widely recognized evaluation metric to comprehensively assess method performance. AU$\overline{\text{PRC}}$ is specifically suited to multi-label classification, providing a global measure of performance across all classes. The metric is computed using micro-averaged precision ($\overline{\text{Prec}}$) and recall ($\overline{\text{Rec}}$), which are calculated as follows:
\begin{equation}
\overline{\text{Prec}} = \frac{\sum_i \text{TP}_i}{\sum_i \text{TP}_i + \sum_i \text{FP}_i},\quad
\overline{\text{Rec}} = \frac{\sum_i \text{TP}_i}{\sum_i \text{TP}_i + \sum_i \text{FN}_i},
\end{equation}
where $\text{TP}_i$, $\text{FP}_i$, and $\text{FN}_i$ denote the true positives, false positives, and false negatives for class $i$, respectively. By varying the decision threshold, a precision-recall curve is generated, and the area beneath this curve provides a single scalar performance measure summarizing precision and recall trade-offs across classes. AU$\overline{\text{PRC}}$ is preferred over accuracy or F1-score for multi-label tasks because: (i) it is threshold-independent, avoiding arbitrary cutoff selection; (ii) it handles class imbalance well by focusing on positive predictions; and (iii) it provides a comprehensive view of precision-recall trade-offs across the operating range.

AU$\overline{\text{PRC}}$ is computed using the established implementation available in the \texttt{scikit-learn} library~\cite{scikit}.

\subsection{Broader Generalizability Assessment}\label{app:generalization}

Table~\ref{tab:hierarchical_comparison_full} shows the complete performance comparison across all hierarchical levels for all nine datasets, extending the results presented in Table~\ref{tab:main_results}. MAPLE* (with semantic initialization) consistently outperforms both MAPLE (with random initialization) and flat baselines across remote sensing, medical imaging, and fine-grained visual categorization domains.

\begin{table}[t!]
\scriptsize
\caption{Complete performance comparison across hierarchical levels using AU$\overline{\textrm{PRC}}$ (\%). This table extends Table~\ref{tab:main_results} from the main paper with results across all nine datasets and all hierarchical levels. $l_i$ denotes the $i$-th hierarchical level. MAPLE* uses semantic initialization while MAPLE uses random initialization. $\Delta^*$ and $\Delta$ show relative improvements over flat baseline.}
\label{tab:hierarchical_comparison_full}
\centering
\begin{tabular}{l|c|cc|c|cc}
\hline
\textbf{Dataset} & & \textbf{MAPLE*} & \textbf{MAPLE} & \textbf{Flat Baseline} & $\boldsymbol{\Delta^*}$ \textbf{(\%)} & $\boldsymbol{\Delta}$ \textbf{(\%)} \\
\hline
\hline
\multicolumn{7}{c}{\cellcolor{green!20}\textit{Remote Sensing Datasets}} \\
\hline
\multirow{4}{*}{AID} & $l_1$ & \textbf{95.31} & 94.11& - & - & - \\
& $l_2$ & \textbf{94.32} & 93.10 & - & - & - \\
& $l_3$ & \textbf{84.83} & 82.50 & - & - & - \\
& Leaf & \textbf{87.25} & 86.66 & 84.21 & +3.61 & +2.91 \\
\hline
\multirow{4}{*}{DFC-15} & $l_1$ & \textbf{99.62} & 99.51& - & - & - \\
& $l_2$ & 98.33 & \textbf{99.41} & - & - & - \\
& $l_3$ & \textbf{98.37} & 98.35 & - & - & - \\
& Leaf & \textbf{98.71} & 98.36 & 98.65 & +0.06 & -0.29 \\
\hline
\multirow{4}{*}{MLRSNet} & $l_1$ & \textbf{98.23} & 98.21 & - & - & - \\
& $l_2$ & \textbf{97.88} & 97.11 & - & - & - \\
& $l_3$ & \textbf{96.77} & 96.50 & - & - & - \\
& Leaf & \textbf{96.71} & 96.31 & 96.17 & +0.56 & +0.15 \\
\hline
\hline
\multicolumn{7}{c}{\cellcolor{blue!20}\textit{Medical Imaging Datasets}} \\
\hline
\multirow{4}{*}{MuRed} & $l_1$ & \textbf{78.69} & 78.60 & - & - & - \\
& $l_2$ & 73.24 & \textbf{73.25} & - & - & - \\
& $l_3$ & \textbf{49.00} & 48.11 & - & - & - \\
& Leaf & \textbf{55.04} & 54.23 & 53.52 & +2.84 & +1.33 \\
\hline
\multirow{2}{*}{HPA} & $l_1$ & \textbf{79.08} & 78.01 & - & - & - \\
& Leaf & \textbf{51.18} & 48.50 & 44.99 & +13.76 & +7.80 \\
\hline
\multirow{6}{*}{PadChest} & $l_1$ & \textbf{72.28} & 72.21 & - & - & - \\
& $l_2$ & \textbf{38.16} & 38.10 & - & - & - \\
& $l_3$ & 15.29 & \textbf{15.30} & - & - & - \\
& $l_4$ & \textbf{9.57} & 9.56 & - & - & - \\
& $l_5$ & \textbf{8.35} & 8.10 & - & - & - \\
& Leaf & \textbf{14.24} & 13.11 & 11.68 & +21.92 & +12.24 \\
\hline
\hline
\multicolumn{7}{c}{\cellcolor{red!20}\textit{Fine-Grained Visual Categorization}} \\
\hline
\multirow{2}{*}{Pets} & $l_1$ & \textbf{99.85} & 99.80 & - & - & - \\
& Leaf & \textbf{93.62} & 92.30  & 91.99 & +1.77 & +0.34 \\
\hline
\multirow{2}{*}{Cars} & $l_1$ & \textbf{97.89} & 96.31 & - & - & - \\
& Leaf & \textbf{93.14} & 92.50 & 84.54 & +10.17 & +9.41 \\
\hline
\multirow{4}{*}{ETHEC} & $l_1$ & \textbf{99.74} & 99.31 & - & - & - \\
& $l_2$ & \textbf{99.22} & 99.10 & - & - & - \\
& $l_3$ & \textbf{96.55} & 93.31 & - & - & - \\
& Leaf & \textbf{86.89} & 81.31 & 78.72 & +10.38 & +3.29 \\
\hline
\end{tabular}
\end{table}

\begin{figure}[t]
   \centering
   \begin{minipage}[b]{0.7\textwidth}
       \centering
       \includegraphics[width=\textwidth]{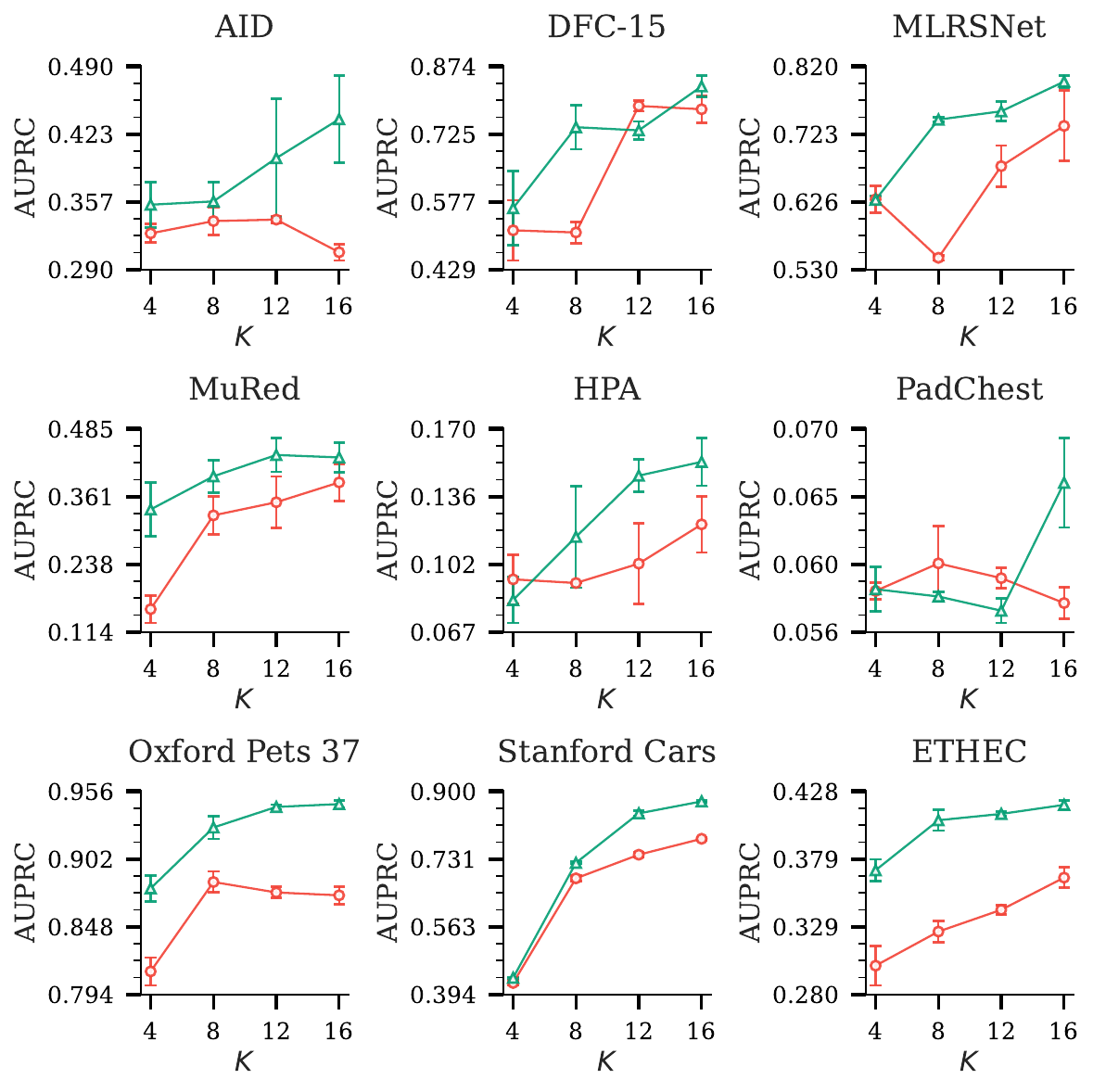}
       {\footnotesize }
   \end{minipage}
   \caption{
   Few-shot learning performance comparison between MAPLE (in \textbf{\textcolor{darkgreen}{green}}) 
   and flat MLC baseline (in \textbf{\textcolor{darkred}{red}}) across representative datasets. 
   Results show AU$\overline{\textrm{PRC}}$ performance ($\mu \pm \sigma$ across three experimental repeats) 
   for varying numbers of shots per category ($K \in \{4, 8, 12, 16\}$). 
   MAPLE consistently outperforms the baseline, with particularly pronounced benefits in low-data regimes.
   }
   \label{fig:few_shot_learning_curves}
\end{figure}

Figure~\ref{fig:few_shot_learning_curves} demonstrates the consistent performance advantages of MAPLE* over the flat MLC baseline across majority of datasets and shot configurations. The hierarchical approach shows particularly pronounced benefits in low-data regimes, with the performance gap being most substantial at $K = 4$ and $K = 8$ shots per category. This pattern indicates that 
taxonomic relationships among classes can compensate for limited direct supervision.

The magnitude of improvement varies significantly across domains and correlates with hierarchical complexity. Medical datasets exhibit the most substantial gains, with HPA and MuRed showing considerable performance differences that persist across all shot settings. Remote sensing datasets demonstrate more modest but consistent improvements, while FGVC benchmarks show intermediate gains with notable variation. Stanford Cars and ETHEC display strong hierarchical benefits that become more pronounced as shot count increases, suggesting that these fine-grained domains particularly benefit from structured learning even with moderate data availability.

Interestingly, some datasets like PadChest show complex performance patterns where the hierarchical advantage varies with shot count, potentially reflecting the interaction between taxonomic structure complexity and available training signal. The consistent upward trend for MAPLE across increasing shot counts demonstrates that hierarchical supervision scales effectively with data availability while maintaining its advantage over the flat MLC approach.

\subsection{Computational Efficiency Analysis}\label{app:efficiency}

Table~\ref{tab:computational_efficiency} presents detailed computational overhead analysis. MAPLE adds only 0.7\% additional GFLOPs and 2.6\% more parameters compared to the flat baseline, with inference time overhead ranging from 2.4\% to 5.3\% across different batch sizes.

\begin{table}[h]
\centering
\caption{Computational efficiency comparison on AID dataset.}
\label{tab:computational_efficiency}
\scriptsize
\begin{NiceTabular}{l|c c c c|c|c}
\toprule
& \multicolumn{4}{c|}{Per Image (ms)} & & \\
\cmidrule(lr){2-5}
Model & $8$ & $16$ & $32$ & $64$ & GFLOPs & Params (M) \\
\midrule
Flat Baseline & 3.02 & 2.96 & 2.62 & 2.55 & 33.54 & 86.57 \\
MAPLE & 3.18 & 3.01 & 2.65 & 2.61 & 33.77 & 88.84 \\
\midrule
Overhead (\%) & +5.3 & +1.7 & +1.1 & +2.4 & +0.7 & +2.6 \\
\bottomrule
\end{NiceTabular}
\end{table}

\subsection{Qualitative Results}\label{sec:qualitative_results}

This section presents a qualitative evaluation of the proposed method on three representative datasets: AID (remote sensing), Oxford Pets (fine-grained visual categorization), and MuRed (medical imaging). Unless stated otherwise, we use the semantically initialized variant (SentenceTransformer-based initialization) and refer to it simply as MAPLE throughout.

\subsubsection{Embedding Evolution Analysis}\label{sec:embedding_evolution}

\begin{figure*}[ht!]
    \centering
    \begin{minipage}[b]{1\textwidth}
        \centering
        \includegraphics[width=\textwidth]{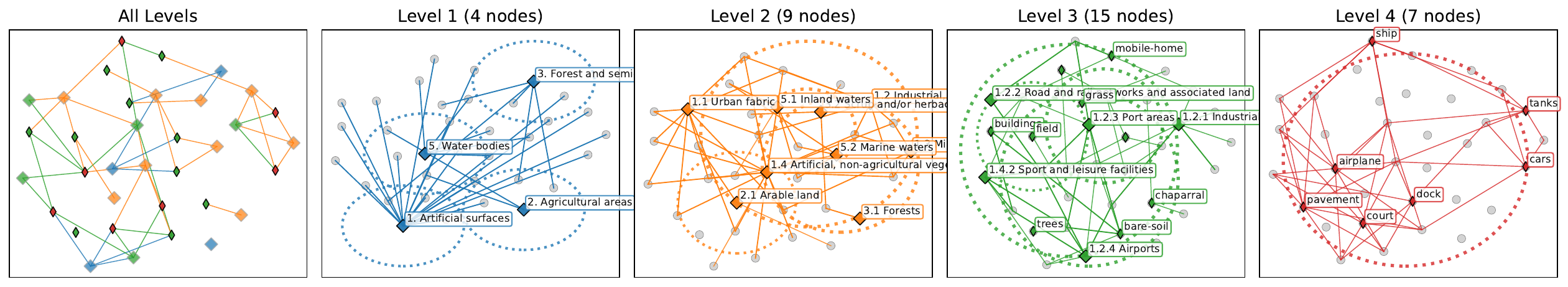}
        {\footnotesize (a) Initial node embeddings}
    \end{minipage}
    
    \vspace{0.5em}
    
    \begin{minipage}[b]{1\textwidth}
        \centering
        \includegraphics[width=\textwidth]{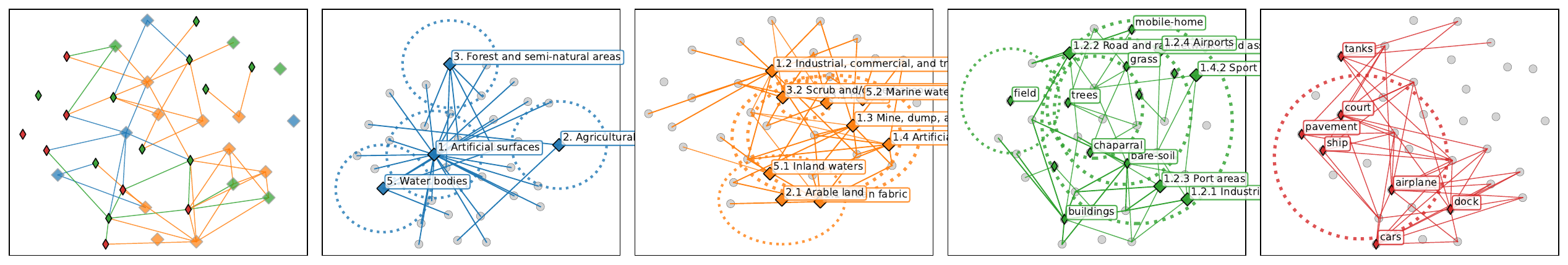}
        {\footnotesize (b) Learned node embeddings}
    \end{minipage}
    
    \vspace{0.5em}
    \begin{minipage}[b]{1\textwidth}
        \centering
        \includegraphics[width=\textwidth]{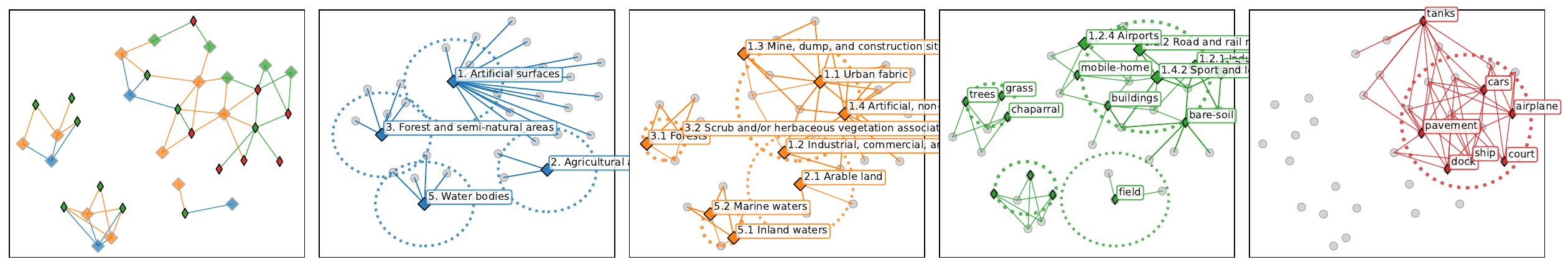}
        {\footnotesize (c) GNN node embeddings}
    \end{minipage}
    
    \vspace{0.5em}
    \begin{minipage}[b]{1\textwidth}
        \centering
        \includegraphics[width=\textwidth]{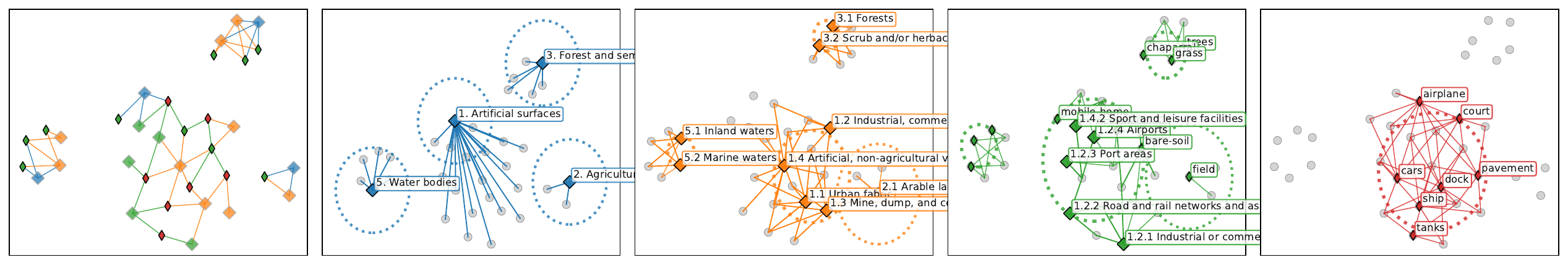}
        {\footnotesize (d) Visual-fused node embeddings}
    \end{minipage}
    
    \vspace{0.5em}
    
    \begin{minipage}[b]{1\textwidth}
        \centering
        \includegraphics[width=\textwidth]{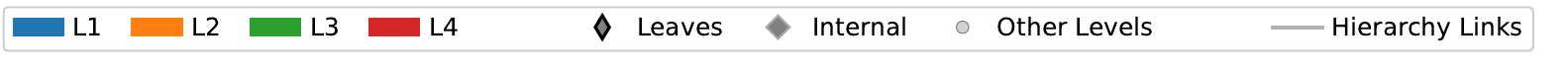}
    \end{minipage}
    
    \caption{Evolution of node embeddings throughout MAPLE training stages visualized using UMAP dimensionality reduction on the AID dataset. (a) Initial embeddings show random spatial distribution. (b) Learned embeddings after hierarchical semantic initialization demonstrate clear semantic clustering. (c) GNN embeddings after graph neural network refinement exhibit enhanced separation between semantic groups. (d) Visual-fused embeddings after multimodal fusion achieve sophisticated organization, integrating both visual and semantic information while respecting hierarchical structure.}
    \label{fig:embedding_evolution}
\end{figure*}

For simplicity and clarity of presentation, we visualize the learned embeddings for the AID dataset as a representative example. Figure~\ref{fig:embedding_evolution} demonstrates the progressive emergence of semantic structure across four key training stages. The initial embeddings (a) exhibit random spatial distribution with no semantic coherence. Following hierarchical semantic initialization (b), embeddings start to develop semantic organization where taxonomically related nodes tend to cluster together. Graph neural network refinement (c) further enhances clustering through iterative message passing, achieving improved separation with enhanced intra-cluster cohesion. The final multimodal fusion (d) produces the most sophisticated organization, where the adaptive gating mechanism integrates visual and semantic information.

\subsubsection{Error Analysis}\label{app:error_analysis}

To understand the practical benefits of hierarchical classification, we conduct a detailed analysis of leaf-level classification errors comparing MAPLE with the baseline MLC method. Table~\ref{tab:confusion_reduction} quantifies these improvements across three representative datasets. MAPLE achieves consistent error reduction, with the most substantial improvement on Oxford Pets (70.7\% reduction). This significant improvement on a fine-grained recognition task demonstrates MAPLE's particular strength in scenarios where hierarchical relationships between classes are semantically meaningful. The AID dataset shows a moderate but consistent improvement (12.1\%), while MuRed exhibits minimal improvement (1.3\%), likely due to its already low baseline confusion rate.

\begin{table}[h]
\scriptsize
\centering
\caption{Leaf-level confusion reduction across datasets. MAPLE consistently reduces confusion instances compared to baseline models, with particularly strong improvements on Oxford Pets (70.7\% reduction) and AID (12.1\% reduction). Lower values indicate better performance.}
\label{tab:confusion_reduction}
\begin{tabular}{lcccc}
\toprule
\textbf{Dataset} & \textbf{Baseline} & \textbf{MAPLE} & \textbf{Improvement (\%)} & \textbf{Absolute Reduction} \\
\midrule
\cellcolor{green!20}\textbf{AID} & 107 & 94 & 12.1 & 13 \\
\cellcolor{blue!20}\textbf{MuRed} & 77 & 76 & 1.3 & 1 \\
\cellcolor{red!20}\textbf{Oxford Pets} & 41 & 12 & 70.7 & 29 \\
\bottomrule
\end{tabular}
\end{table}

\begin{figure}[t]
   \centering
   \begin{minipage}[b]{1\textwidth}
       \centering
       \includegraphics[width=\textwidth]{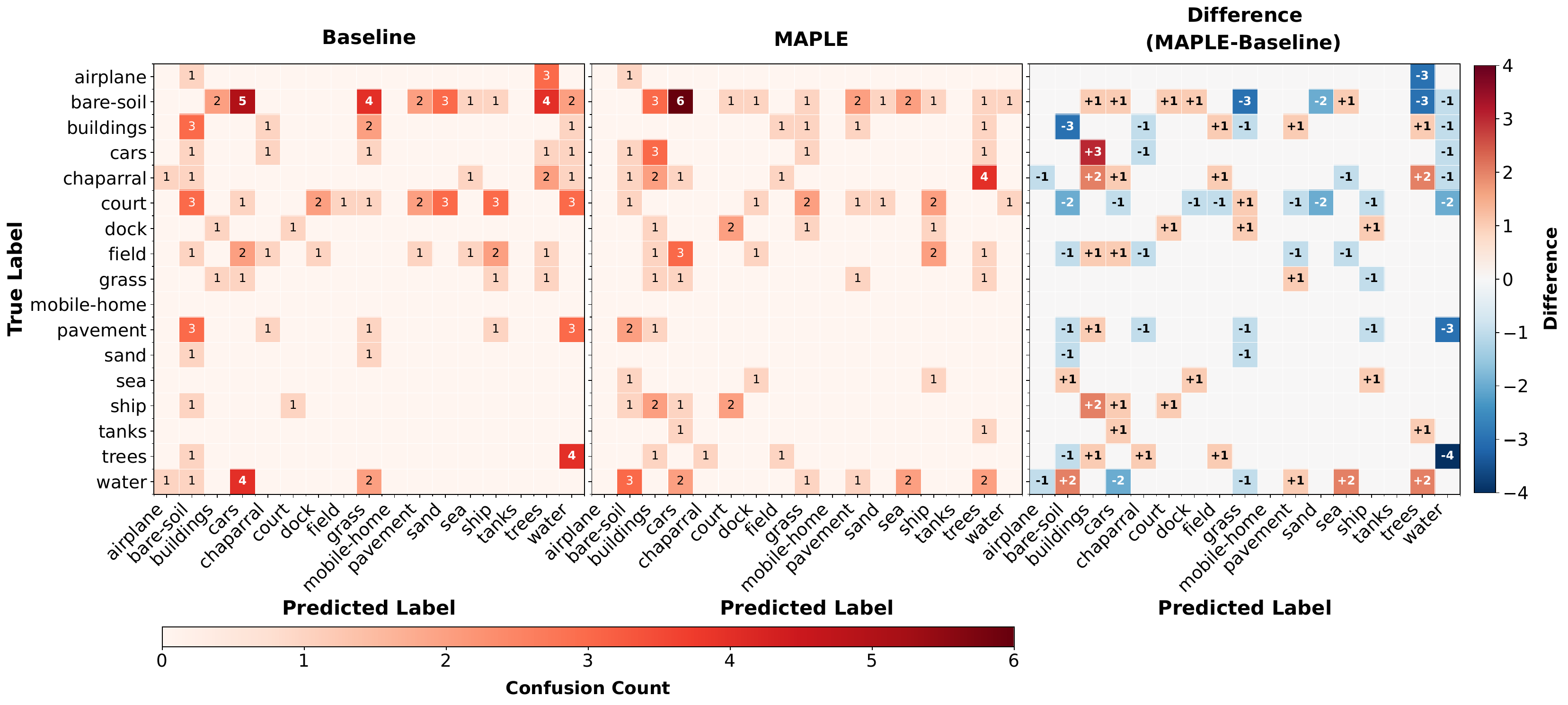}
   \end{minipage}
   \caption{Leaf-level confusion matrix comparison for AID dataset. Left: baseline model confusions, Center: MAPLE confusions, Right: difference matrix (MAPLE $-$ Baseline). Blue cells indicate reduced confusions (improvements), red cells indicate increased confusions. MAPLE achieves a 12.1\% reduction in total leaf-level confusions, with notable improvements in semantically challenging pairs such as trees$\rightarrow$water and buildings$\rightarrow$bare-soil.}
   \label{fig:leaf_comparison}
\end{figure}

For a detailed understanding of these improvements, we present a comprehensive confusion matrix analysis for the AID dataset as a representative example. Figure~\ref{fig:leaf_comparison} shows confusion matrices of the baseline MLC and MAPLE, and their difference matrix. The difference matrix clearly illustrates where MAPLE reduces errors (blue cells) versus where new errors emerge (red cells).

The most significant improvements occur in semantically challenging confusion pairs. MAPLE substantially reduces problematic confusions such as trees-to-water and pavement-to-water misclassifications, suggesting that hierarchical representations help the model better distinguish between natural and artificial surface types. Similarly, reductions in structural versus natural category confusions indicate improved understanding of taxonomic relationships.

The error analysis reveals that MAPLE's hierarchical structure provides the most benefit when classes have clear taxonomic relationships, visual similarities can be resolved through higher-level semantic understanding, and sufficient hierarchical structure exists to guide the learning process. These findings validate our hypothesis that incorporating hierarchical knowledge leads to more robust and semantically coherent predictions at the leaf level.

\subsection{Ablation Studies}\label{app:ablation}

\subsubsection{Graph Encoder Analysis}

\begin{figure}[t]
   \centering
   \includegraphics[width=1\textwidth]{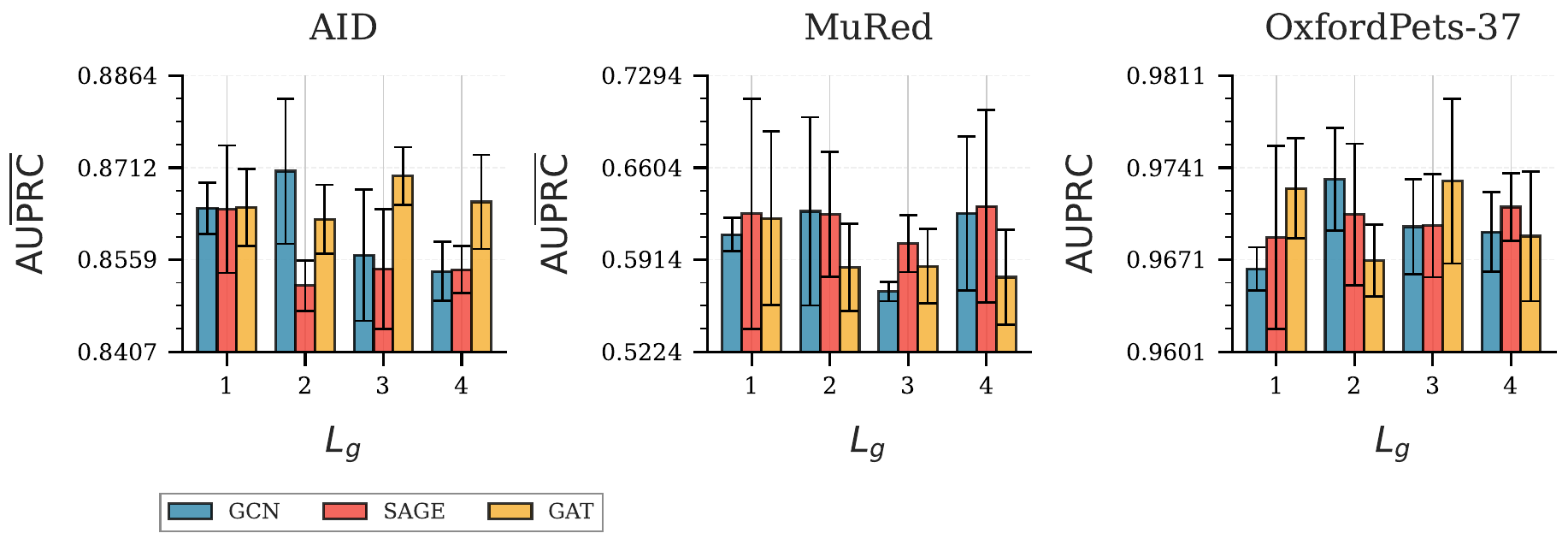}
   \caption{Performance comparison of different GNN architectures across datasets.}
   \label{fig:gnn_comparison}
\end{figure}

We analyze the impact of different graph neural network architectures on MAPLE's performance by comparing three widely-used GNN variants: Graph Convolutional Networks (GCN)\cite{kipf2016semi}, GraphSAGE\cite{hamilton2017inductive}, and Graph Attention Networks (GAT)\cite{velivckovic2017graph}. Figure~\ref{fig:gnn_comparison} shows the performance across different numbers of GNN layers ($L_g = 1, 2, 3, 4$) on three datasets.

The results demonstrate that all three GNN architectures achieve comparable performance, with minor variations across datasets. GCN shows consistent performance across layer depths, while GAT exhibits slight improvements with deeper architectures on AID and Oxford Pets-37. GraphSAGE performs competitively but shows more sensitivity to layer depth on certain datasets. Notably, the performance differences between architectures are relatively small (typically within 0.01 AUPRC), suggesting that the hierarchical message passing mechanism is more important than the specific aggregation strategy.

The optimal number of layers varies by dataset, with most configurations benefiting from 2-3 layers. Deeper networks ($L_g = 4$) sometimes show diminishing returns, indicating that the hierarchical structure can be effectively captured with moderate network depth.

\subsubsection{Embedding Initialization Strategies}

The initialization of node embeddings in hierarchical classification systems plays a crucial role in model convergence and final performance. While random initialization has been the standard approach in many graph-based models, recent advances in sentence embedding models offer opportunities to leverage semantic priors that can better capture the inherent relationships between class labels. Recent work~\cite{koloski2025llmembeddingsdeeplearning} has shown that different LLM embeddings can improve deep learning architectures for training on downstream tasks.

We evaluate five initialization strategies across representative datasets from different domains: (1) Random initialization with normalized Gaussian vectors; (2) NV-Embed-v2 initialization using the state-of-the-art generalist embedding model\cite{nvembedv2}; (3) MPNet-Base-v2 initialization using the all-mpnet-base-v2 sentence transformer\cite{mpnet}; (4) Word2Vec initialization using pre-trained word vectors\cite{ord2vec}; and (5) GloVe initialization with pre-trained embeddings\cite{glove}. For semantic methods, we enhance class names with hierarchical context by incorporating parent-child relationships from the taxonomy structure (See Appendix~\ref{app:implementation}). All experiments are repeated three times with different random seeds to ensure statistical reliability.

\begin{figure}[t]
   \centering
   \includegraphics[width=1\textwidth]{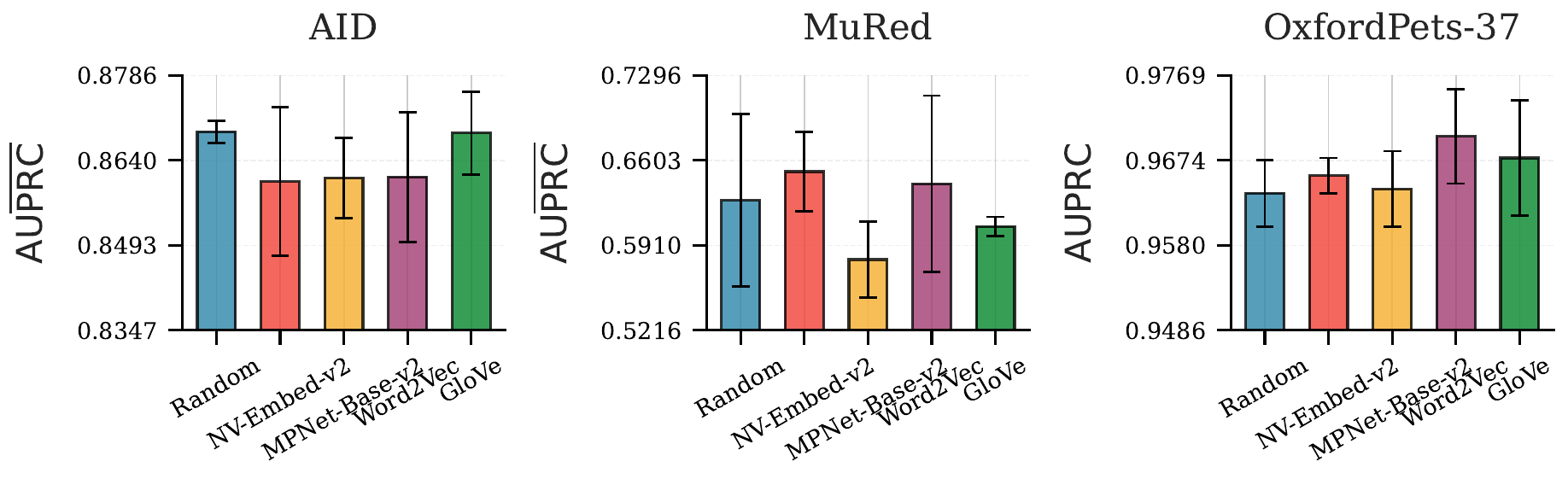}
   \caption{Comparison of embedding initialization strategies.}
   \label{fig:different_embeddings}
\end{figure}

Figure~\ref{fig:different_embeddings} presents the comparative results across three representative datasets. Surprisingly, random initialization consistently achieves competitive or superior performance compared to sophisticated pre-trained embedding strategies. On AID, random initialization (0.8665C
AUPRC) slightly outperforms all semantic alternatives, while on MuRed, it achieves comparable results to the best semantic method. Only on Oxford Pets-37 do semantic methods show marginal improvements, with Word2Vec achieving the highest performance (0.9683 AUPRC).

These results reveal several important insights about hierarchical visual classification. Random initialization provides maximum plasticity, allowing the model to learn task-specific visual relationships without semantic bias. The strong visual features extracted by the ViT backbone appear sufficient for discovering meaningful hierarchical relationships through graph-based propagation. In contrast, pre-trained semantic embeddings may impose text-corpus relationships that do not align with visual similarities, potentially constraining the model's ability to learn optimal visual hierarchies.

The modest performance differences between methods and relatively small standard deviations across repeats suggest that the GNN architecture effectively refines initial embeddings regardless of initialization strategy. This finding indicates that our hierarchical approach successfully learns visual taxonomies from data, with the graph-based refinement process being more influential than semantic priors for final performance.

\subsection{Limitations}\label{app:limitations}

While MAPLE demonstrates strong performance across diverse HMLC tasks, several areas present opportunities for future improvement.

MAPLE's performance is influenced by the quality of the input hierarchical structure. Our hybrid construction approach combines expert-curated taxonomies (CORINE, ICD-10) with ChatGPT-generated hierarchies, which may occasionally introduce inconsistencies. Language model-generated taxonomies, while useful, may not always fully capture domain-specific relationships or could reflect biases present in training data, particularly in specialized domains such as medical imaging or remote sensing.

While MAPLE effectively models fine-grained distinctions between leaf categories, UMAP visualizations occasionally show that semantically related categories sharing common parent nodes may appear separated in the learned embedding space. This suggests there is room for improvement in fully exploiting higher-level taxonomic relationships across all hierarchy levels.

The benefits of hierarchical modeling vary across different domains. While medical imaging datasets consistently show substantial improvements, some remote sensing datasets (e.g., MLRSNet) exhibit more modest gains despite their large scale and complex structures. This variation indicates that the effectiveness of hierarchical modeling may depend on specific domain characteristics.

\clearpage
\section*{EurIPS Paper Checklist}

\begin{enumerate}

\item {\bf Claims}
    \item[] Question: Do the main claims made in the abstract and introduction accurately reflect the paper's contributions and scope?
    \item[] Answer: \answerYes{}
    \item[] Justification: The abstract and Introduction describe MAPLE's components (hierarchical semantic initialization, graph-based structure encoding, adaptive fusion and level-aware objective) and claim consistent gains with small parameter overhead. These claims are supported by Tables~\ref{tab:main_results}–\ref{tab:sota_remote}, the few-shot results in Table~\ref{tab:fewshot}, and analyses in Appendices~\ref{app:generalization}–\ref{app:ablation}.

\item {\bf Limitations}
    \item[] Question: Does the paper discuss the limitations of the work performed by the authors?
    \item[] Answer: \answerYes{}
    \item[] Justification: The Discussion and Appendix~\ref{app:limitations} describe that gains narrow on very large datasets and that performance depends on the quality of the constructed hierarchy; we also note sensitivity to noisy mappings and oversmoothing with deeper GNNs. Computational overheads and latency are quantified in Appendix~\ref{app:efficiency} with details in Table~\ref{tab:computational_efficiency}.

\item {\bf Theory assumptions and proofs}
    \item[] Question: For each theoretical result, does the paper provide the full set of assumptions and a complete (and correct) proof?
    \item[] Answer: \answerNA{}
    \item[] Justification: The paper is empirical and methodological; it does not introduce formal theorems or proofs.

\item {\bf Experimental result reproducibility}
    \item[] Question: Does the paper fully disclose all the information needed to reproduce the main experimental results of the paper to the extent that it affects the main claims and/or conclusions of the paper (regardless of whether the code and data are provided or not)?
    \item[] Answer: \answerYes{}
    \item[] Justification: Datasets, hierarchy construction, model variants, training setup, metrics, and evaluation protocols are detailed in Appendices~\ref{app:datasets}–\ref{app:implementation} and \ref{app:evaluation}. We disclose hardware and compute in Appendix~\ref{app:resources} and efficiency in Appendix~\ref{app:efficiency}. The exact hierarchical label structures will be released as YAML configuration files with the code.

\item {\bf Open access to data and code}
    \item[] Question: Does the paper provide open access to the data and code, with sufficient instructions to faithfully reproduce the main experimental results, as described in supplemental material?
    \item[] Answer: \answerNo{}
    \item[] Justification: All datasets are public and cited, but an anonymized code repository is not included in the submission. We will release code, YAML hierarchies, and run scripts with detailed instructions after the review period, preserving anonymity at submission time.

\item {\bf Experimental setting/details}
    \item[] Question: Does the paper specify all the training and test details (e.g., data splits, hyperparameters, how they were chosen, type of optimizer, etc.) necessary to understand the results?
    \item[] Answer: \answerYes{}
    \item[] Justification: Implementation details, augmentations, optimizer, schedules, batch sizes, epochs, and split strategies are specified in Appendix~\ref{app:implementation} and \ref{app:evaluation}; dataset statistics appear in Table~\ref{tab:datasets}.

\item {\bf Experiment statistical significance}
    \item[] Question: Does the paper report error bars suitably and correctly defined or other appropriate information about the statistical significance of the experiments?
    \item[] Answer: \answerYes{}
    \item[] Justification: We report averages over three runs for all settings and show mean$\pm$std in the learning curves (Fig.~\ref{fig:few_shot_learning_curves}) and in the few-shot table (Table~\ref{tab:fewshot}). For brevity, other tables report means only; variability is visible in the curves.

\item {\bf Experiments compute resources}
    \item[] Question: For each experiment, does the paper provide sufficient information on the computer resources (type of compute workers, memory, time of execution) needed to reproduce the experiments?
    \item[] Answer: \answerYes{}
    \item[] Justification: Hardware is specified in Appendix~\ref{app:resources} (4$\times$A100 40\,GB). Appendix~\ref{app:efficiency} provides inference-time, GFLOPs, and parameter overhead (Table~\ref{tab:computational_efficiency}), with consistent scaling across batch sizes.

\item {\bf Code of ethics}
    \item[] Question: Does the research conducted in the paper conform, in every respect, with the NeurIPS Code of Ethics \url{https://neurips.cc/public/EthicsGuidelines}?
    \item[] Answer: \answerYes{}
    \item[] Justification: We evaluate on public datasets with appropriate citations, do not process personally identifiable information, and adhere to standard research practices and anonymization.

\item {\bf Broader impacts}
    \item[] Question: Does the paper discuss both potential positive societal impacts and negative societal impacts of the work performed?
    \item[] Answer: \answerNo{}
    \item[] Justification: While applications and potential benefits for environmental monitoring, urban planning, and medical analysis are discussed in the Introduction and Discussion, a dedicated broader impacts section with explicit negative impact analysis is not included.

\item {\bf Safeguards}
    \item[] Question: Does the paper describe safeguards that have been put in place for responsible release of data or models that have a high risk for misuse (e.g., pretrained language models, image generators, or scraped datasets)?
    \item[] Answer: \answerNA{}
    \item[] Justification: The work does not release high-risk models or scraped web datasets; we train task models on curated public benchmarks.

\item {\bf Licenses for existing assets}
    \item[] Question: Are the creators or original owners of assets (e.g., code, data, models), used in the paper, properly credited and are the license and terms of use explicitly mentioned and properly respected?
    \item[] Answer: \answerNo{}
    \item[] Justification: We cite all datasets and prior methods, but specific license names are not listed in the current version. We will include license details for each asset in the camera-ready version.

\item {\bf New assets}
    \item[] Question: Are new assets introduced in the paper well documented and is the documentation provided alongside the assets?
    \item[] Answer: \answerNA{}
    \item[] Justification: No new datasets are introduced at submission time. Configuration files (YAML hierarchies) and trained weights will be documented and released with the code after the review period.

\item {\bf Crowdsourcing and research with human subjects}
    \item[] Question: For crowdsourcing experiments and research with human subjects, does the paper include the full text of instructions given to participants and screenshots, if applicable, as well as details about compensation (if any)? 
    \item[] Answer: \answerNA{}
    \item[] Justification: The research does not involve crowdsourcing or human subjects.

\item {\bf Institutional review board (IRB) approvals or equivalent for research with human subjects}
    \item[] Question: Does the paper describe potential risks incurred by study participants, whether such risks were disclosed to the subjects, and whether Institutional Review Board (IRB) approvals (or an equivalent approval/review based on the requirements of your country or institution) were obtained?
    \item[] Answer: \answerNA{}
    \item[] Justification: The research does not involve human subjects.

\item {\bf Declaration of LLM usage}
    \item[] Question: Does the paper describe the usage of LLMs if it is an important, original, or non-standard component of the core methods in this research? Note that if the LLM is used only for writing, editing, or formatting purposes and does not impact the core methodology, scientific rigorousness, or originality of the research, declaration is not required.
    \item[] Answer: \answerYes{}
    \item[] Justification: We used ChatGPT for limited assistance in hierarchy mapping when direct alignment to established taxonomies was ambiguous, followed by manual verification (Appendix~\ref{app:datasets}). We also used it for light paraphrasing to improve clarity. LLMs were not used to design methods, run experiments, or interpret results.
\end{enumerate}

\end{document}